\documentclass[letterpaper]{article} 
\usepackage{aaai2026}  
\usepackage{times}  
\usepackage{helvet}  
\usepackage{courier}  
\usepackage[hyphens]{url}  
\usepackage{graphicx} 
\usepackage{natbib}  
\usepackage{caption} 
\usepackage{algorithm}
\usepackage{algorithmic}

\usepackage{amsfonts}
\usepackage{tabularx}
\usepackage{amsmath}
\usepackage{multirow}
\usepackage{booktabs}
\usepackage{graphicx}
\usepackage{amssymb}
\usepackage{multirow}
\usepackage{subcaption} 

\usepackage{newfloat}
\usepackage{listings}
\DeclareCaptionStyle{ruled}{labelfont=normalfont,labelsep=colon,strut=off} 
\floatstyle{ruled}
\newfloat{listing}{tb}{lst}{}
\floatname{listing}{Listing}
\title{The Semantic Architect: How FEAML Bridges Structured Data and LLMs for Multi-Label Tasks}
\author {
    Wanfu Gao\textsuperscript{\rm 1,\rm 2},
    Zebin He\textsuperscript{\rm 1,\rm 2},
    Jun Gao\textsuperscript{\rm 1,\rm 2}\thanks{Corresponding author}
}
\affiliations {
    \textsuperscript{\rm 1}College of Computer Science and Technology, Jilin University, China\\
    \textsuperscript{\rm 2}Key Laboratory of Symbolic Computation and Knowledge Engineering of Ministry of Education, Jilin
University, China\\
    gaowf@jlu.edu.cn, hezb2121@mails.jlu.edu.cn, gaocheng23@mails.jlu.edu.cn
}

\begin{document}

\maketitle

\begin{abstract}
Existing feature engineering methods based on large language models (LLMs) have not yet been applied to multi-label learning tasks. They lack the ability to model complex label dependencies and are not specifically adapted to the characteristics of multi-label tasks. To address the above issues, we propose Feature Engineering Automation for Multi-Label Learning (FEAML), an automated feature engineering method for multi-label classification which leverages the code generation capabilities of LLMs. By utilizing metadata and label co-occurrence matrices, LLMs are guided to understand the relationships between data features and task objectives, based on which high-quality features are generated. The newly generated features are evaluated in terms of model accuracy to assess their effectiveness, while Pearson correlation coefficients are used to detect redundancy. FEAML further incorporates the evaluation results as feedback to drive LLMs to continuously optimize code generation in subsequent iterations. By integrating LLMs with a feedback mechanism, FEAML realizes an efficient, interpretable and self-improving feature engineering paradigm. Empirical results on various multi-label datasets demonstrate that our FEAML outperforms other feature engineering methods.
\end{abstract}

\begin{links}
    \link{Datasets}{https://github.com/hezb2121/FEAML-Datasets}
\end{links}

\section{Introduction}

\begin{figure}[ht]
\centering
\includegraphics[width=0.95\columnwidth]{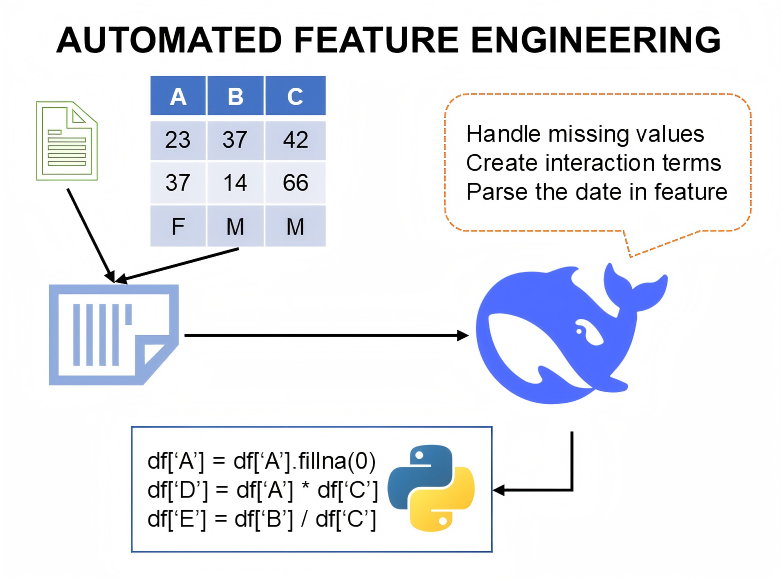}
\caption{Automated Feature Engineering Workflow. Natural language requests are translated into code via LLMs, enabling automatic preprocessing and feature generation.}
\label{fig_1}
\end{figure}

In recent years, large language models (LLMs) have demonstrated remarkable general capabilities across various machine learning tasks, particularly in the field of feature engineering~\cite{han2024large,mann2020language,zhao2023survey}.
 Their powerful semantic understanding and code generation abilities make them valuable tools for automating feature processing~\cite{dinh2022lift}. As illustrated in Figure~\ref{fig_1}, existing research has begun exploring the use of LLMs for feature construction and selection on structured data, showing promising results in single-label classification and regression tasks~\cite{hollmann2023caafe}. However, most of these studies focus on learning scenarios involving a single output variable and systematic exploration in the more complex domain of multi-label classification remains limited.

Multi-label classification is widely present in real-world applications such as image annotation~\cite{mao2025implicit,hao2024label}, text classification~\cite{hang2024dual} and information retrieval~\cite{yang2025fast}, where each instance is typically associated with multiple labels~\cite{zhou2025batch}. Compared to traditional single-label classification, multi-label tasks are more challenging due to complex inter-label dependencies, a combinatorially large label space and limited training data, often leading to model overfitting~\cite{liu2015large,mao2024learning,han2025multi}. As a result, feature engineering becomes especially critical in multi-label scenarios. Nevertheless, traditional feature engineering methods typically rely on manual design and fixed templates, which are insufficient to address the requirements.

To address the issue in multi-label feature engineering, we propose Feature Engineering Automation for Multi-Label Learning (FEAML). To the best of our knowledge, FEAML is the first method to systematically introduce LLMs into this domain. FEAML adopts a prompt-driven, closed-loop design for feature generation and evaluation. The prompt construction is guided by two complementary sources: (1) structured metadata extracted from the dataset, such as feature types, missing rates and label distributions; (2) a label co-occurrence matrix that captures inter-label dependencies. The above information is translated into natural language prompts and input into the LLMs to generate Python code for feature construction.

FEAML statically validates the generated code for safety before executing it dynamically via exec function to produce new features, which are then merged with the original dataset. Lightweight models are used to decide whether to retain each new feature for multi-label classification. To avoid redundancy and dimensional explosion, FEAML also computes the Pearson correlation ($\rho$) between new and original features, discarding any feature with $|\rho| > 0.95$. Through this closed-loop process of generation, evaluation and selection, FEAML constructs an efficient, robust and generalizable feature space that significantly enhances downstream multi-label learning performance.

In summary, the principal contributions of this work are outlined as follows:
\begin{itemize}
    \item A pioneering and systematic integration of LLMs into the domain of multi-label feature engineering, effectively addressing a long-standing research gap and laying the groundwork for future advancements in automated feature construction.
    \item The development of standardized multi-label adaptations for a suite of representative structured datasets, thereby furnishing a comprehensive and reusable benchmark resource to facilitate rigorous empirical studies in multi-label learning.
    \item Extensive empirical evaluation of the proposed FEAML method, which consistently surpasses conventional methodologies across a range of tasks and performance metrics, demonstrating not only superior predictive capability but also robust generalization and strong practical utility.
\end{itemize}

\begin{figure*}[t]
\centering
\includegraphics[width=0.8\textwidth]{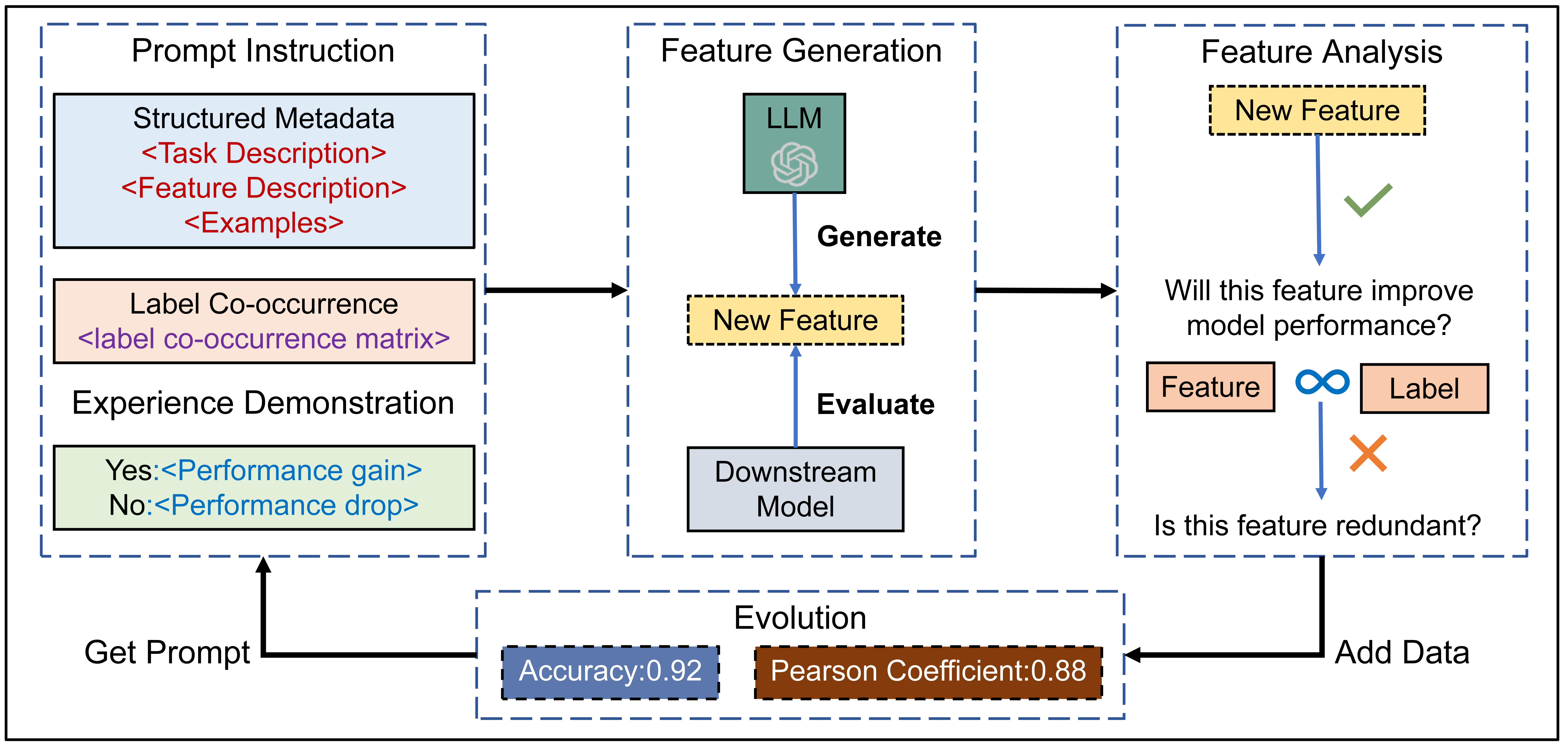} 
\caption{An overview of the proposed FEAML method. The method integrates structured metadata and label co-occurrence statistics to construct context-aware prompts, which guides LLMs in generating transformation code for automated feature engineering. FEAML includes modules for metadata extraction, prompt generation, code verification, feature evaluation and feedback-driven optimization, forming a closed-loop pipeline for multi-label classification tasks.}
\label{fig_2}
\end{figure*}

\section{Related work}

\subsection{Multi-label Learning}
Multi-label learning tasks face a significant challenge due to their large output space~\cite{gouk2016learning,han2025enhanced}. Unlike traditional single-label classification, multi-label learning allows each instance to be associated with multiple labels simultaneously, resulting in an exponential increase in the number of possible label combinations~\cite{gong2023distributed,jia2023learning}. Given a total of \(L\) labels, the potential label combination space size is \(2^L\). This combinatorial explosion makes it difficult for training data to cover all possible label sets, causing severe data sparsity issues~\cite{liu2025oversampling,hao2024anchor}. To address the above issues, leveraging label correlations has become a widely adopted strategy to improve learning effectiveness. Generally, existing methods can be grouped into three categories based on the order of label dependencies they consider: first-order, second-order and high-order approaches. First-order methods treat each label independently~\cite{boutell2004learning,zhang2018binary}. Second-order methods model pairwise interactions between labels~\cite{brinker2014graded,hao2024exploring}. High-order methods capture dependencies among subsets or the entire set of labels~\cite{huang2019supervised,read2011classifier}.

FEAML belongs to the category of high-order methods. It constructs a label co-occurrence matrix and leverages its statistical patterns to generate structured prompts, which guide LLMs in generating task-relevant features. In addition, FEAML computes the Pearson correlation coefficients between the newly generated features and the labels to assess their discriminative capability across different label dimensions. By doing so, FEAML effectively captures the underlying relationships among labels, thereby enhancing both the discriminability and diversity of the generated features.

\subsection{Feature Engineering with LLMs}
Recent studies have increasingly investigated the use of LLMs to automate various stages of feature engineering. While these efforts demonstrate the potential of LLMs in this domain, the current body of research primarily emphasizes feasibility exploration rather than the development of comprehensive, scalable methodologies. The applications of LLMs are largely concentrated in rule extraction and feature generation. FeatLLM~\cite{han2024large} introduces a pipeline in which LLMs are employed to extract feature construction rules, which are subsequently applied to linear regression tasks. Although this approach illustrates a viable path toward automated feature engineering, the generated features are predominantly binary and specifically designed for linear models, thereby limiting their applicability in broader machine learning scenarios. CAAFE~\cite{hollmann2023caafe} adopts a more straightforward approach by combining prompt-based guidance with iterative interactions between LLMs and a tabular prediction model. This semi-automated framework simplifies the feature engineering process and lowers the entry barrier for non-expert users. However, the effectiveness of the resulting features is highly dependent on prompt quality and lacks systematic optimization. ELLM-FT~\cite{gong2025evolutionary} explores the integration of LLMs with reinforcement learning techniques to guide feature expansion. In this method, LLMs serve as policy advisors, suggesting strategies for evolving feature representations. Despite its potential, this method involves complex policy design and incurs significant computational overhead. OCTree~\cite{nam2024optimized} proposes a decision-tree-inspired approach that first generates candidate feature names and then derives transformation rules accordingly. This method enhances the relevance and effectiveness of feature generation by streamlining the reasoning process and has demonstrated strong performance in task-specific scenarios.

In summary, existing LLMs-based feature engineering methods have focused exclusively on single-label tasks and have not addressed the challenges unique to multi-label learning, such as complex label dependencies and the need for more expressive feature representations. FEAML is proposed to address this underexplored problem. It not only leverages LLMs to generate high-quality and interpretable feature code but also utilizes label co-occurrence matrices and feature-label correlation analysis to guide feature generation and selection, making it better suited for multi-label tasks.

\section{Our Proposed Method}
\subsection{Problem Definition}

Let $\mathcal{D} = \{(x_i, y_i)\}_{i=1}^N$ denote the multi-label training dataset, where $x_i \in \mathbb{R}^S$ is the input feature vector of sample $i$, and $y_i \in \{0,1\}^L$ is the associated binary label vector. Each $x_i$ can also be interpreted as an ordered set of $S$ elements representing structured attributes. The label vector $y_i = [y_{i1}, y_{i2}, \ldots, y_{iL}]$ consists of $L$ binary values, where $y_{ij} = 1$ indicates that the $j$-th label is relevant to sample $i$, and $y_{ij} = 0$ indicates irrelevance. The objective of multi-label learning is to learn a function $f: \mathbb{R}^S \rightarrow \{0,1\}^L$ that accurately predicts the relevant label set for an unseen instance.

\subsection{Overall Framework}

As shown in Figure~\ref{fig_2}, FEAML integrates structured metadata and label co-occurrence statistics to construct context-aware natural language prompts, which are used to guide LLMs in performing automated feature engineering. Specifically, the method first extracts structured metadata from the original dataset, including feature names, data types, missing rates and statistical properties. These metadata are converted into natural language templates that provide field-level semantic and statistical context, enabling LLMs to generate appropriate transformation code, such as normalization and discretization. In parallel, FEAML constructs a label co-occurrence matrix to quantify joint frequencies and conditional probabilities among label pairs, revealing potential inter-label dependencies. This statistical information is further embedded into high-level reasoning prompts to instruct the LLMs in generating interaction-based or ratio-based features that reflect label dependencies.

Once all prompts are constructed, they are fed into the LLMs, which automatically generates structured and executable Python code to perform various feature transformation and derivation operations. To ensure runtime safety and stability, the generated code undergoes static analysis and sandbox verification, allowing only safe and compliant functions to be executed. The verified code is then run to produce candidate features.

These candidate features are merged with the original features to form an expanded feature space. FEAML employs lightweight models such as Random Forests and XGBoost to evaluate each new feature, assessing its impact on performance metrics such as Accuracy and Hamming Loss. Only features that yield measurable performance improvements are retained, while others are discarded.

To continuously enhance feature generation quality, FEAML incorporates a closed-loop feedback mechanism that dynamically adjusts prompting strategies based on evaluation results. Ineffective prompts are penalized, and template formats are optimized to improve the relevance and diversity of subsequent prompts.

In summary, FEAML establishes a fully automated, end-to-end feature engineering pipeline—from metadata extraction, prompt construction, and code generation, to feature evaluation and feedback optimization. This pipeline significantly reduces manual effort and demonstrates strong stability and modeling efficiency in multi-label classification tasks.

\begin{figure}[ht]
\centering
\includegraphics[width=0.95\columnwidth]{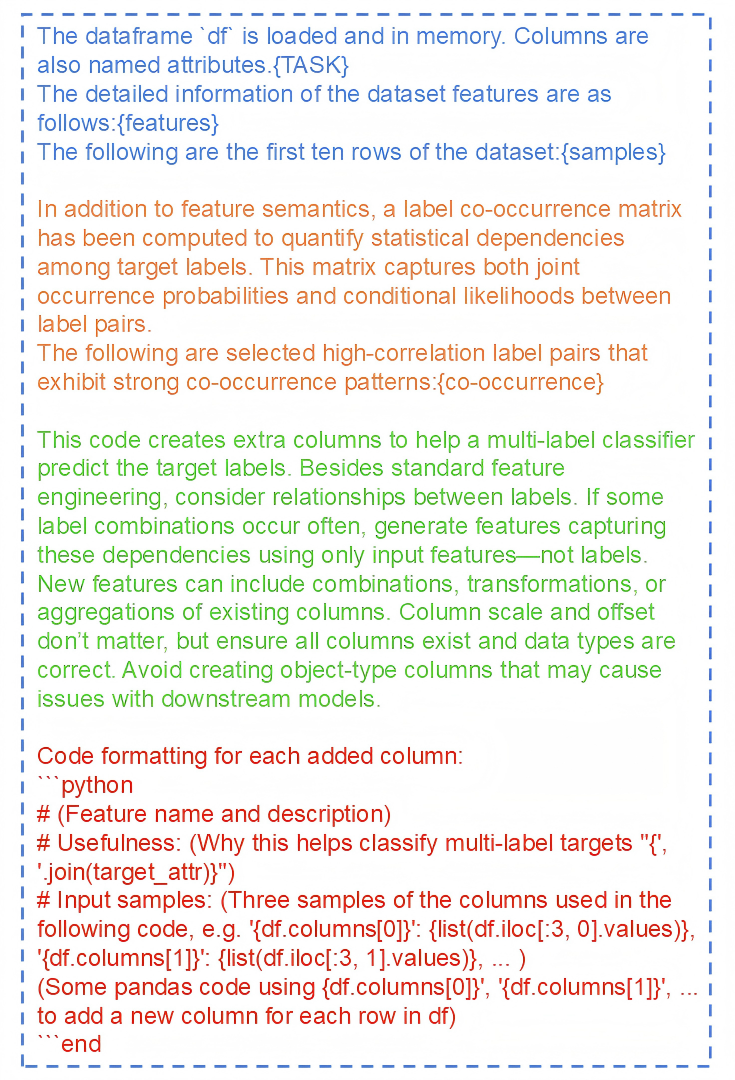}
\caption{Prompts used for feature generation: the blue text indicates structured meta-information prompts; the orange text represents label co-occurrence matrix prompts; the green text is designed to enhance LLMs reasoning; the red text guides structured output.
}
\label{fig_3}
\end{figure}

\subsection{Prompt Design for Feature Generation}
\paragraph{Structured Metadata Prompts.}
After obtaining the multi-label dataset, we construct a standardized structured metadata prompt to assist LLMs in understanding the task objectives and data structure. As shown in the blue text in Figure~\ref{fig_3}, this prompt consists of three core components: task description, feature descriptions and raw data examples. 
First, the task description part concisely summarizes the task goal in natural language and clearly states that it is a multi-label classification task. For example, taking the bank dataset as an example, the task is described as follows: Predict multiple financial-related attributes such as whether the client subscribed to a term deposit, has a housing loan or a personal loan, using customer profile data from telemarketing campaigns. Next, the feature descriptions provide a structured summary for each input attribute, including field names, data types and statistical properties. For categorical fields, their value ranges and frequencies are also provided. For instance, the feature famsize takes values LE3 and GT3, with proportions of 33.5\% and 66.5\%, respectively. 
Finally, we directly embed the first ten rows of the dataset in tabular form. Each row contains complete feature values along with the corresponding label information. This structured presentation enables the LLMs to intuitively observe the actual data distribution and label annotation patterns, thereby facilitating a better understanding of the task requirements.

\paragraph{Label Co-occurrence Prompts.}

An essential component of the FEAML method is the use of label co-occurrence statistics to guide the construction of semantically meaningful prompts for the LLMs. To capture inter-label dependencies, we compute a label co-occurrence matrix that quantifies the joint frequency and conditional probabilities of label pairs across the training data. Specifically, given a binary label matrix $\mathbf{Y} \in \{0, 1\}^{n \times L}$, where $n$ is the number of instances and $L$ is the number of labels, the label co-occurrence matrix $\mathbf{C} \in \mathbb{R}^{L \times L}$ is defined as:

\begin{equation}
\mathbf{C}_{ij} = \frac{1}{n} \sum_{k=1}^n \mathbb{I}[Y_{ki} = 1 \land Y_{kj} = 1],
\end{equation}
where $\mathbf{C}_{ij}$ represents the empirical joint probability that both label $i$ and label $j$ appear in the same instance. $\mathbb{I}[\cdot]$ is the indicator function. Additionally, we compute the conditional probability:

\begin{equation}
P(j \mid i) = \frac{\mathbf{C}_{ij}}{\sum_{k=1}^n \mathbb{I}[Y_{ki} = 1]},
\end{equation}
where $P(j \mid i)$ denotes the probability of label $j$ being present given that label $i$ is active.

After computing the label co-occurrence matrix, we incorporate the co-occurrence and conditional probability information as label correlation prompts for the LLMs. It can be found in the orange-colored text in Figure~\ref{fig_3}. This provides the LLMs with a semantic context of the global label structure, enriching the depth and hierarchy of the prompt information.

\paragraph{Reasoning Part.}
As shown in the green section of Figure~\ref{fig_3}, this prompt clearly guides the model to generate additional features that are useful for multi-label classification, with a particular emphasis on considering relationships between labels while strictly avoiding the use of the labels themselves to prevent data leakage. It encourages extracting valuable information from input features through combinations, transformations and aggregations to enhance model performance. The prompt also pays attention to feature usability by avoiding the creation of object-type columns that may cause issues for downstream models and relaxes restrictions on numerical scale, making feature construction more flexible and practical.

\paragraph{Standardized Output.}
As shown in the red part of Figure~\ref{fig_3}, this structured output format explains how to generate code for each newly added feature. It consists of three parts: first, the feature name and description; second, an explanation of how the feature contributes to classifying the multi-label targets; finally, three sample values from the input columns used, along with the corresponding code snippet that generates a new feature column for each row of the dataset based on these inputs.

\subsection{Code Generation and  Execution}

In the code execution phase, the generated Python code strings are dynamically executed using the exec function. Prior to execution, the code undergoes static safety checks to ensure the absence of syntax errors and unsafe calls. After execution via exec, the new features are added to the DataFrame, thereby realizing the feature generation. This process allows flexible execution of diverse feature transformation code produced by LLMs, while ensuring runtime safety and data integrity.

\subsection{Feature Evaluation and Selection}

After executing the generated Python code to obtain new features, FEAML merges them with the original features to form an expanded feature space. To assess the practical value of these newly generated features, the method incorporates a systematic evaluation and selection mechanism to ensure that only features with genuine predictive contribution are retained while redundant ones are filtered out.

During the evaluation phase, FEAML employs lightweight models, such as Random Forests, to assess the utility of each candidate feature. Specifically, for each new feature \( f \), it is appended to the original feature set \( \mathbf{X} \) to form an expanded feature set \( \mathbf{X}^+ = [\mathbf{X}, f] \) and a model is trained using cross-validation. The method then evaluates the impact of including this feature on standard multi-label classification metrics using Accuracy (\(\text{Acc}\)) and Hamming Loss (\(\text{HL}\)). If the feature \( f \) leads to an improvement in at least one core metric:

\begin{equation}
\Delta \text{Acc} > 0 \quad \text{or} \quad \Delta \text{HL} < 0,
\end{equation}
it is marked as a potential candidate for retention.

However, performance improvement alone does not guarantee feature quality. To prevent overfitting caused by dimensional explosion, FEAML further integrates a redundancy detection mechanism to analyze the overlap between newly generated features and existing ones. This mechanism primarily relies on Pearson correlation.

Let the new feature be \( f \) and any original feature be \( x_i \). The Pearson correlation coefficient \( \rho(f, x_i) \) is calculated as:

\begin{equation}
\rho(f, x_i) = \frac{\mathrm{cov}(f, x_i)}{\sigma_f \sigma_{x_i}},
\end{equation}
where \(\mathrm{cov}(\cdot, \cdot)\) denotes covariance and \(\sigma\) denotes standard deviation. If there exists any \( i \) such that

\begin{equation}
|\rho(f, x_i)| > 0.95,
\end{equation}
then \( f \) is considered redundant and discarded.

By combining predictive performance evaluation and redundancy analysis, FEAML can filter a large pool of generated features and retain only those that provide real value and structural complementarity. These selected features not only significantly improve predictive performance but also demonstrate stable behavior across different data splits and label configurations.

\subsection{Downstream Models Prediction}
After multiple rounds of iterative feature generation, FEAML filters out high-quality features from a large pool of candidates—those that demonstrate strong predictive power, structural complementarity, and minimal redundancy. These selected features are then combined with the original input features to form an enhanced feature set, which serves as the input for training downstream multi-label classification models. This feature fusion approach effectively enriches the model's input representation and improves its ability to capture complex label correlations.

\begin{table*}[ht]
\centering
\begin{tabular}{lccccccc}
\toprule
\toprule

\multicolumn{8}{c}{\textbf{Datasets}} \\
\cmidrule{2-8}
\textbf{Methods} & adult & bank & communities & credit-g & heart & myocardial & student  \\
\cmidrule{2-8}
\multicolumn{8}{c}{\textbf{Accuracy} $\uparrow$} \\
\midrule
Base & 0.6183 & 0.6114 & 0.5238 & 0.4900 & 0.5018 & 0.1304 & 0.1772  \\
LDA& 0.6187 & 0.6141 & 0.5241 & 0.4878 & 0.5031 & 0.1205 & 0.1796  \\
AFAT & 0.6188 & 0.6156 & 0.5240 & 0.4901 & 0.5065 & 0.1265 & 0.1799  \\

GRFG & 0.6201 & 0.6194 & 0.5298 & 0.4955 & 0.5097 & 0.1331 & 0.1845  \\
CAAFE& 0.6251 & 0.6255 & 0.5301 & 0.4952 & 0.5101 & \textbf{0.1395} & 0.1887  \\
FeatLLM & 0.6274 & 0.6214 & 0.5344 & 0.5012 & 0.5094 & 0.1294 & 0.1954  \\

\textbf{FEAML(Ours)} & \textbf{0.6312} & \textbf{0.6319} & \textbf{0.5478} & \textbf{0.5101} & \textbf{0.5201} & 0.1355 & \textbf{0.1962}  \\

\midrule
\multicolumn{8}{c}{\textbf{Hamming Loss} $\downarrow$} \\
\midrule
Base & 0.2726 & 0.2135 & 0.0685 & 0.0983 & 0.1766 & 0.1299 & 0.2151  \\
LDA& 0.2720 & 0.2127 & 0.1246 & 0.0980 & 0.1756 & 0.1301 & 0.2149  \\
AFAT & 0.2719 & 0.2128 & 0.0791 & 0.0980 & 0.1742 & 0.1302 & 0.2142  \\

GRFG & 0.2439 & 0.2145 & 0.0653 & 0.0951 & 0.1759 & 0.1288 & 0.2196  \\
CAAFE& 0.2456 & 0.2398 & 0.0651 & 0.0946 & 0.1743 & 0.1294 & 0.2188  \\
FeatLLM & 0.2533 & 0.2067 & 0.0644 & 0.0956 & 0.1798 &\textbf{0.1275} & 0.2174  \\

\textbf{FEAML(Ours)} & \textbf{0.2385} & \textbf{0.2021} & \textbf{0.0632} & \textbf{0.0888} & \textbf{0.1740} & 0.1277 &\textbf{ 0.2094}  \\

\midrule
\multicolumn{8}{c}{\textbf{F1-score(micro)} $\uparrow$} \\
\midrule
Base & 0.8690 & 0.6367 & 0.9279 & 0.8613 & 0.8759 & 0.6788 & 0.8039  \\
LDA& 0.8688 & 0.6368 & 0.9281 & 0.8645 & 0.8761 & 0.6881 & 0.8123  \\
AFAT & 0.8691 & 0.6371 & 0.9281 & 0.8753 & 0.8766 & 0.6871 & 0.8234  \\

GRFG & 0.8692 & 0.6377 & 0.9231 & 0.8759 & 0.8712 & 0.6854 & 0.8345  \\
CAAFE& 0.8720 & 0.6213 & 0.9256 & 0.8812 & 0.8777 & 0.6813 & 0.8254  \\
FeatLLM & 0.8723 & 0.6534 & 0.9271 & 0.8832 & 0.8823 & 0.6743 & 0.8134  \\

\textbf{FEAML(Ours)} & \textbf{0.8867} & \textbf{0.6724} & \textbf{0.9286} & \textbf{0.8954} & \textbf{0.8923} & \textbf{0.6978} &\textbf{0.8477}  \\

\bottomrule
\bottomrule
\end{tabular}
\caption{Performance comparison of different feature generation methods across multiple multi-label datasets. Results are evaluated using Accuracy (higher is better), Hamming Loss (lower is better) and F1-score (higher is better). FEAML consistently achieves competitive performance, demonstrating the effectiveness of LLMs-assisted feature generation.}

\label{tab_1} 
\end{table*}

\begin{table}[ht]
\centering
\renewcommand{\arraystretch}{1.2}
\begin{tabular}{lccc}
\toprule
\toprule
\textbf{Dataset} & \textbf{\#Samples} & \textbf{\#Features} & \textbf{\#Labels} \\
\midrule
adult & 48,842 & 11 & 4 \\
bank & 45,211 & 13 & 2 \\
communities & 1,994 & 94 & 9 \\
credit-g & 1,000 & 19 & 6 \\
heart & 918 & 10 & 2 \\
myocardial & 686 & 85 & 16 \\
student & 395 & 26 & 7 \\

\bottomrule
\bottomrule
\end{tabular}
\caption{Summary of datasets used in multi-label classification}
\label{tab:datasets}
\end{table}

\section{Experiments}

\subsection{Experimental Setup}
\paragraph{Datasets}
To evaluate the generalization ability and robustness of our proposed FEAML method in real-world multi-label scenarios, we systematically adapt seven publicly available structured datasets. These datasets are originally designed for single-label classification tasks. We transform them into multi-label classification tasks. The transformation process incorporates domain knowledge and the relationships between features and labels, enabling semantic expansion and reconstruction of the label space to ensure both task relevance and learning complexity.

These datasets are collected from platforms including Kaggle~\cite{howard2024kaggle}, UCI database~\cite{public2024} and OpenML~\cite{public2024openml}, covering various real-world domains like social behavior, finance and healthcare. Table~\ref{tab:datasets} summarizes the key characteristics of these datasets and the number of constructed labels. Specifically, for each dataset, we identify the primary target variable and constructed auxiliary labels based on features that are semantically related. For example, in social behavior datasets such as adult and student, we introduce labels such as ``marital status'', ``high income'' and ``foreign residence'' to simulate user profiling scenarios with multi-dimensional attributes. In finance and healthcare datasets such as credit-g and heart, we construct labels related to credit risk or clinical symptoms to capture the label dependencies commonly found in real-world decision systems.

For datasets containing continuous target variables, we adopt discretization strategies like quantile binning and unsupervised clustering to convert them into distinguishable categorical labels. To ensure the presence of meaningful multi-label relationships, we comput label co-occurrence probabilities to filter representative and non-redundant label combinations. Additionally, we apply resampling and label smoothing techniques to alleviate class imbalance and improve label coverage.

\paragraph{Evaluation Metrics}

We employ three widely used evaluation metrics: Multi-label Accuracy, Hamming Loss and F1-score. They together provide a comprehensive view of the model’s predictive capabilities. Detailed definitions can be found in~\cite{zhang2013review}.

\paragraph{Baseline Methods}
We compare our method with 6 widely used feature generation methods, including random generation and feature dimension reduction methods: 
(1) \textbf{Base}: using the original dataset without feature generation. (2) \textbf{LDA}~\cite{blei2003latent}, which learns new features through matrix factorization. (3) \textbf{AFAT}~\cite{horn2019autofeat}, which iteratively generates new features and leverages multi-step feature selection to pick useful features. (4) \textbf{GRFG}~\cite{wang2023reinforcement}, which leverages cascading agents and feature group crossing to generate new features. (5) \textbf{CAAFE}~\cite{hollmann2023caafe}, which constructs features based on linguistic context using large language models. (6) \textbf{FeatLLM}~\cite{han2024large}, which enhances large language models' contextual learning through few-shot learning.

\subsection{Overall Performance}

Table~\ref{tab_1} presents the performance comparison of various feature generation strategies across multiple multi-label datasets. The experimental results show that features generated by LLMs significantly enhance the predictive performance, particularly in capturing complex label dependencies. Among all methods, our proposed method FEAML, consistently achieves superior results on 6 datasets, demonstrating strong stability and generalization ability. Compared to existing methods, FEAML exhibits better representation capability in terms of Accuracy, improved predictive power measured by F1-score, with fewer prediction errors reflected by Hamming Loss, confirming its effectiveness in improving multi-label classification performance.

In summary, the LLMs-driven feature generation strategy adopted by FEAML can effectively extract implicit semantic information from structured data, offering a more robust and generalizable solution for multi-label learning tasks.

\subsection{Ablation Study}

To evaluate the contributions of different components in the proposed FEAML method, we conduct ablation experiments on four representative datasets: adult, bank, communities and credit-g. Specifically, we compare the following variants:

\begin{itemize}
    \item \textbf{w/o Structured Metadata Prompts:} Removes prompts containing statistical and semantic information about features. The LLMs can only generate features based on raw data without field-specific context. This variant is referred to as "FEAML+".
    \item \textbf{w/o Label Co-occurrence Prompts:} Disables the use of label co-occurrence statistics in prompt construction, preventing the LLMs from capturing label dependencies in feature generation. This variant is referred to as "FEAML-".
    \item \textbf{Full FEAML (Ours):} The complete framework that integrates all modules including structured prompts and label-aware reasoning.
\end{itemize}

\begin{figure}[ht]
    \centering
    \begin{subfigure}[b]{0.232\textwidth}
        \centering
        \includegraphics[width=\textwidth]{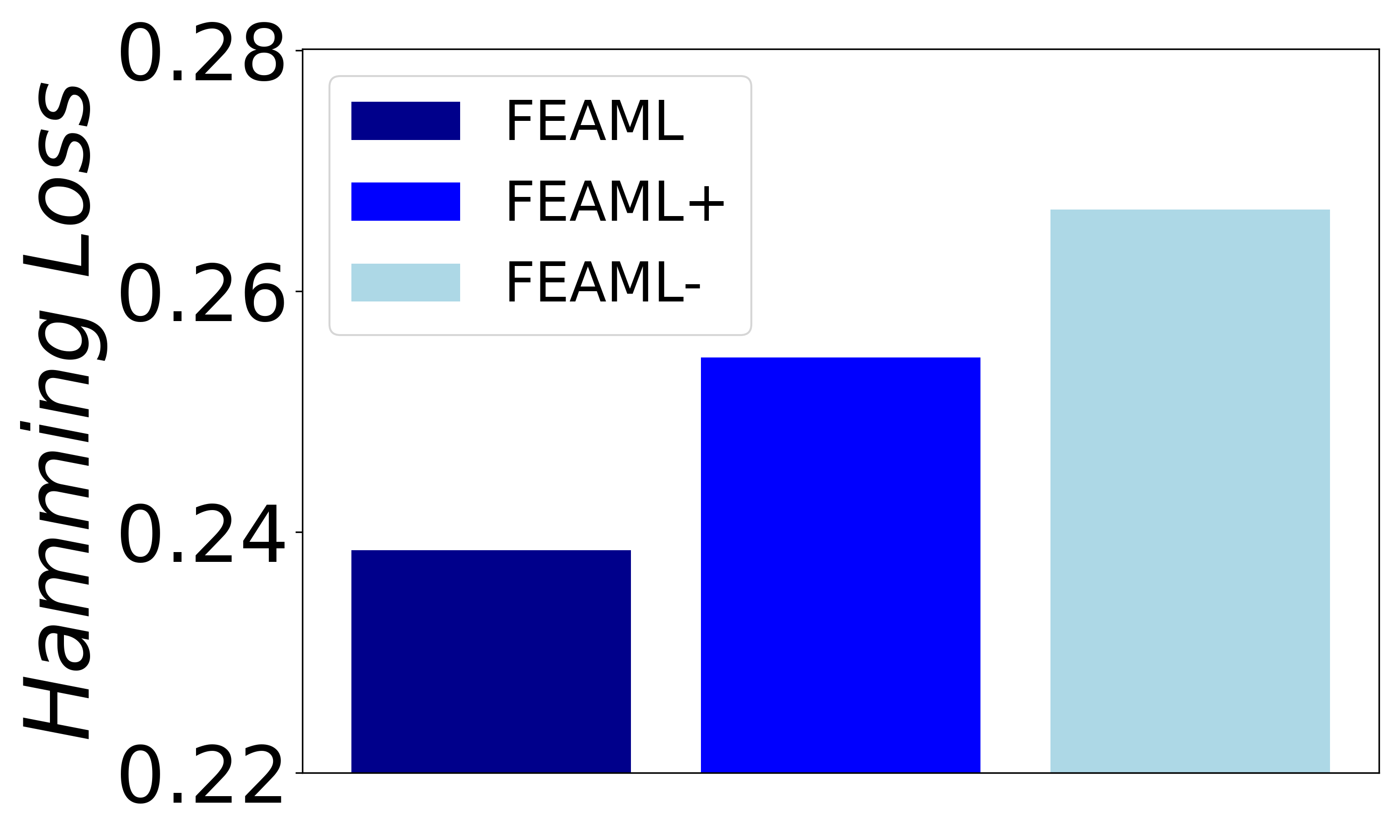}
        \caption{adult}
        \label{fig:adult}
    \end{subfigure}
    \hfill
    \begin{subfigure}[b]{0.232\textwidth}
        \centering
        \includegraphics[width=\textwidth]{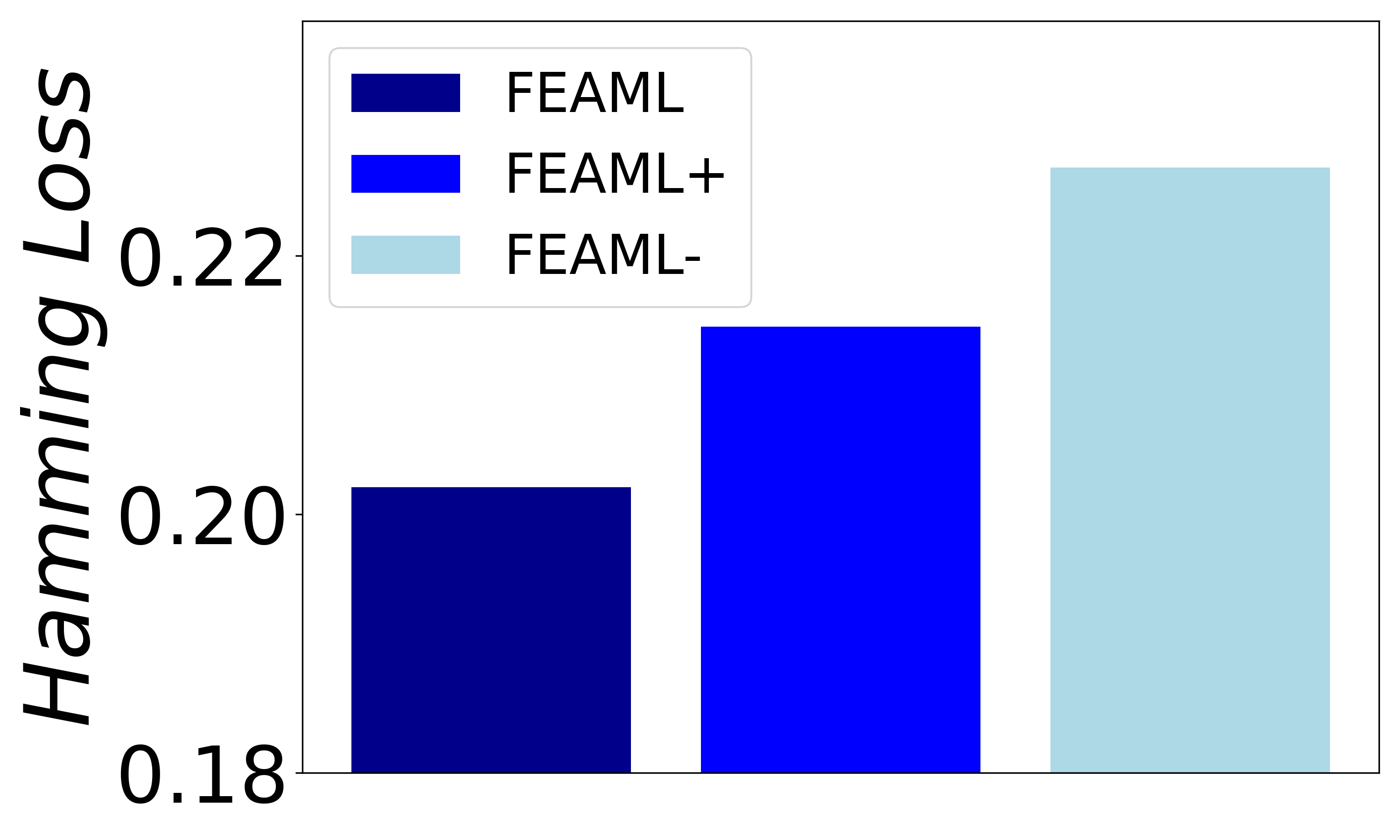}
        \caption{bank}
        \label{fig:bank}
    \end{subfigure}

    \vspace{0.5em} 
    \begin{subfigure}[b]{0.232\textwidth}
        \centering
        \includegraphics[width=\textwidth]{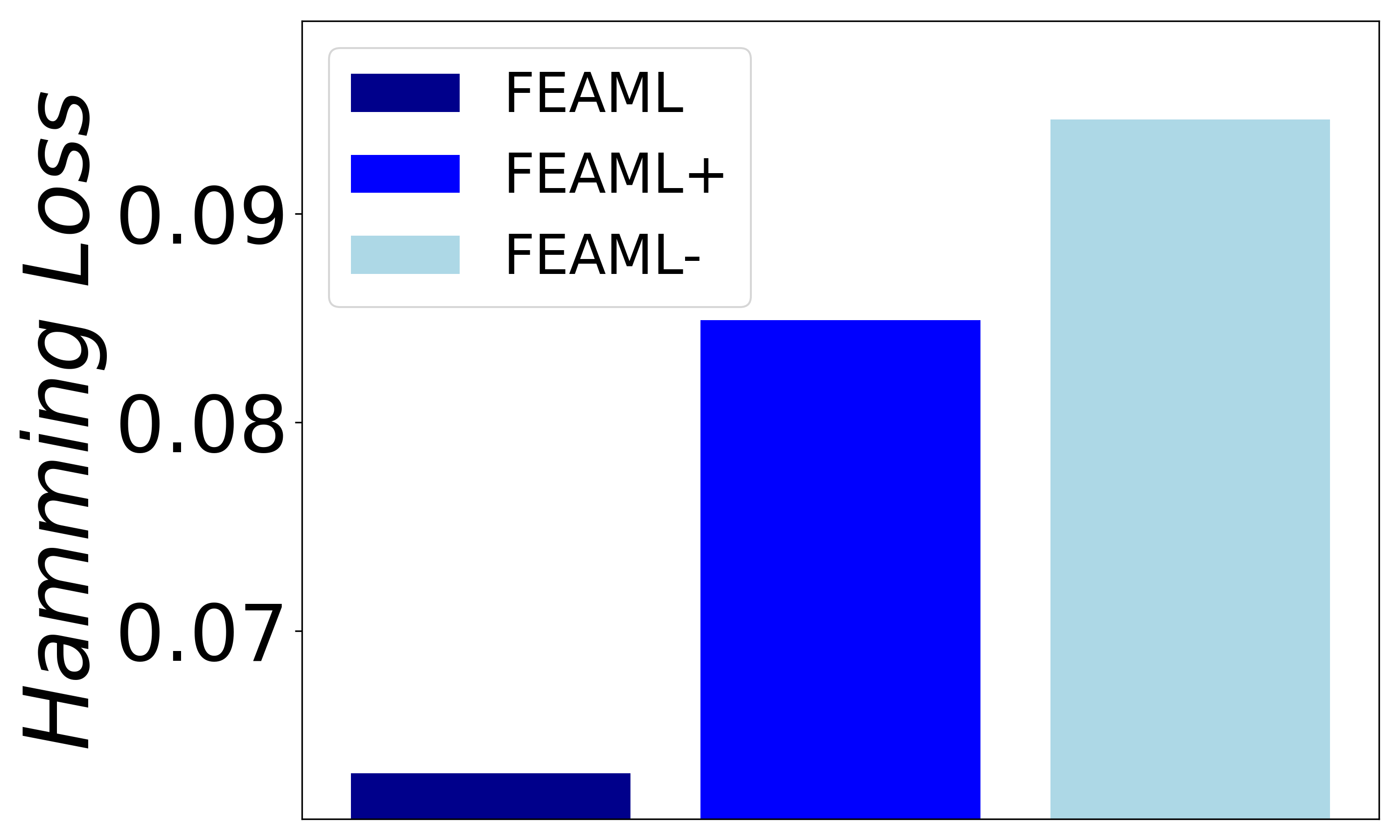}
        \caption{communities}
        \label{fig:communities}
    \end{subfigure}
    \hfill
    \begin{subfigure}[b]{0.232\textwidth}
        \centering
        \includegraphics[width=\textwidth]{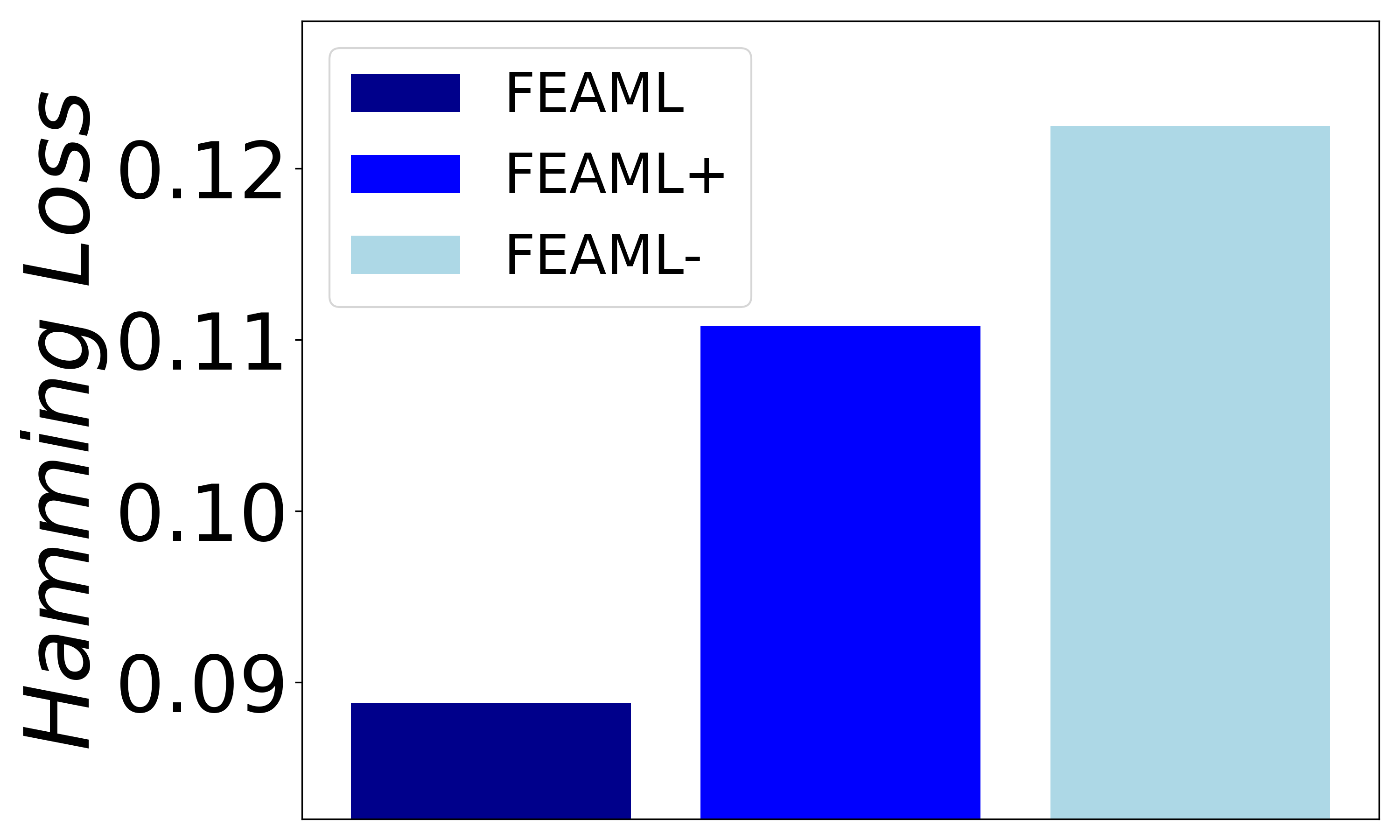}
        \caption{credit-g}
        \label{fig:credit-g}
    \end{subfigure}
    
    \caption{Results of ablation studies on different datasets.}
    \label{fig:ablation}
\end{figure}

We report Hamming Loss using XGBoost as the downstream classifier. Figure~\ref{fig:ablation} summarizes the performance under each setting.

The experimental results clearly indicate that both structured metadata and label co-occurrence prompts significantly enhance model performance. The absence of either component leads to performance degradation, confirming their complementary roles in multi-label feature engineering. Further analysis reveals that the omission of label co-occurrence prompts results in a more pronounced increase in Hamming Loss compared to the absence of structured metadata, demonstrating that label co-occurrence prompts play a more critical role in improving model performance.

\subsection{Comparison on Different Downstream Tasks}

To evaluate the adaptability and effectiveness of the proposed method, we compare its performance across different downstream machine learning models, including Random Forest, XGBoost, Support Vector Machines (SVM), and MLKNN~\cite{zhang2007ml}. The experiment is conducted on the adult dataset.
As shown in Table~\ref{tab:performance}, the proposed method performs well across different downstream models. This highlights the method's ability to generate high-quality features whose effectiveness is not dependent on the complexity of the task or the choice of downstream model. Furthermore, it demonstrates the method's adaptability to various modeling requirements and its potential for broad applications in the field of feature generation.

\begin{table}[h]
\centering
\begin{tabular}{l|cccc}
\toprule
 Methods& RF & XGB & SVM & MLKNN \\
\midrule
Base & 0.6062 & 0.6276 & 0.5829 & 0.5469 \\
LDA & 0.6115 & 0.6478 & 0.6059 & 0.6387 \\
AFAT & 0.5945 & 0.6520 & 0.5603 & 0.6335 \\
CAAFE & 0.6233 & 0.6430 & 0.5970 & 0.6115 \\
FEAML & \textbf{0.6602} & \textbf{0.6530} & \textbf{0.6275} & \textbf{0.6693} \\
\bottomrule
\end{tabular}
\caption{Performance comparison of different downstream models.}
\label{tab:performance}
\end{table}

\begin{figure}[ht]
\centering
\includegraphics[width=0.95\columnwidth]{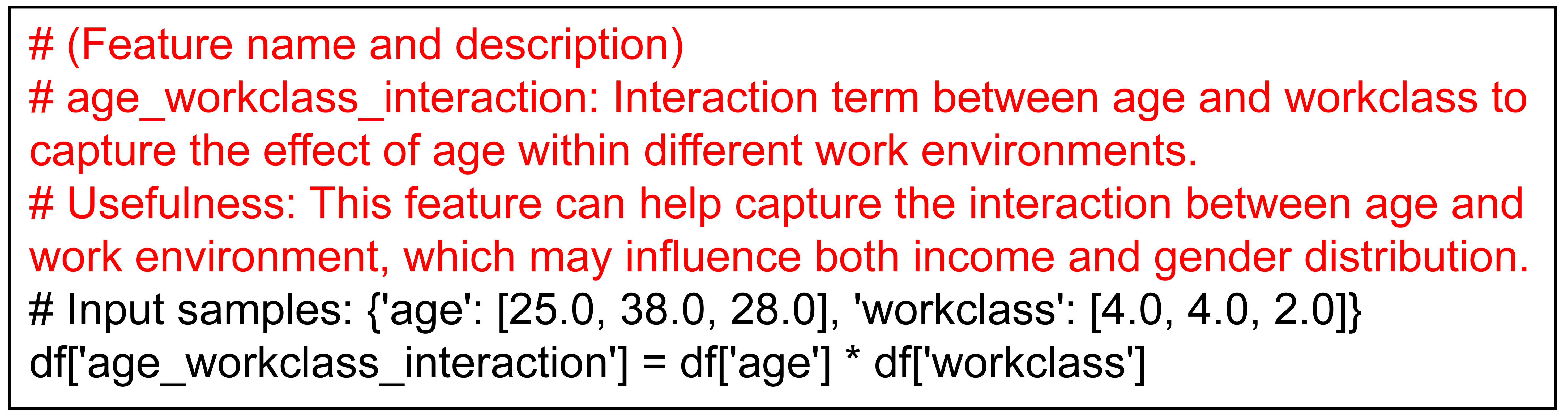}
\caption{Feature age\_workclass\_interaction automatically generated by FEAML.}
\label{fig_5}
\end{figure}

\subsection{Semantic Interpretability}

FEAML generates features that align with human cognitive patterns and offer strong semantic interpretability. Unlike traditional exhaustive transformation methods, FEAML uses structured prompts to guide LLMs in producing features with clear semantic meanings. For example, as shown in Figure~\ref{fig_5}, the feature age\_workclass\_interaction is constructed from the interaction between age and workclass, reflecting the different social implications of age in various occupational contexts.
From a cognitive perspective, a 38-year-old manager and a 38-year-old laborer represent very different socioeconomic statuses. This interaction feature captures such differences and provides the model with a more detailed representation.
By combining metadata and label dependencies, FEAML ensures that generated features are both predictive and interpretable, making it suitable for applications requiring high explainability.

\section{Conclusion}

In this paper, we propose FEAML, an automated feature engineering method for multi-label classification. It generates prompts based on structured metadata and label co-occurrence statistics, and uses LLMs to automatically produce feature transformation code. The method dynamically generates interpretable and task-relevant features, improving the performance of models. Experimental results show that FEAML outperforms traditional methods and other LLMs-based variants in both accuracy and feature quality. This work provides an efficient and scalable solution for applying LLMs to structured data modeling.

\section{Acknowledgments}
This work was supported by the Science Foundation of Jilin 
Province of China under Grant YDZJ202501ZYTS286 and 
in part by Changchun Science and Technology Bureau Project 
under Grant 23YQ05.

\bibliography{aaai2026}

@inproceedings{hollmann2023caafe,
  title={CAAFE: Combining Large Language Models with Tabular Predictors for Semi-Automated Data Science},
  author={Hollmann, Noah and M{\"u}ller, Samuel and Hutter, Frank},
  booktitle={1st Workshop on the Synergy of Scientific and Machine Learning Modeling@ ICML2023},
  year={2023}
}

@article{han2024large,
  title={Large language models can automatically engineer features for few-shot tabular learning},
  author={Han, Sungwon and Yoon, Jinsung and Arik, Sercan O and Pfister, Tomas},
  journal={arXiv preprint arXiv:2404.09491},
  year={2024}
}

@article{boutell2004learning,
  title={Learning multi-label scene classification},
  author={Boutell, Matthew R and Luo, Jiebo and Shen, Xipeng and Brown, Christopher M},
  journal={Pattern recognition},
  volume={37},
  number={9},
  pages={1757--1771},
  year={2004},
  publisher={Elsevier}
}

@article{zhang2018binary,
  title={Binary relevance for multi-label learning: an overview},
  author={Zhang, Min-Ling and Li, Yu-Kun and Liu, Xu-Ying and Geng, Xin},
  journal={Frontiers of Computer Science},
  volume={12},
  pages={191--202},
  year={2018},
  publisher={Springer}
}

@inproceedings{brinker2014graded,
  title={Graded multilabel classification by pairwise comparisons},
  author={Brinker, Christian and Menc{\'\i}a, Eneldo Loza and F{\"u}rnkranz, Johannes},
  booktitle={2014 IEEE International Conference on Data Mining},
  pages={731--736},
  year={2014},
  organization={IEEE}
}

@article{huang2019supervised,
  title={Supervised representation learning for multi-label classification},
  author={Huang, Ming and Zhuang, Fuzhen and Zhang, Xiao and Ao, Xiang and Niu, Zhengyu and Zhang, Min-Ling and He, Qing},
  journal={Machine Learning},
  volume={108},
  pages={747--763},
  year={2019},
  publisher={Springer}
}

@article{read2011classifier,
  title={Classifier chains for multi-label classification},
  author={Read, Jesse and Pfahringer, Bernhard and Holmes, Geoff and Frank, Eibe},
  journal={Machine learning},
  volume={85},
  pages={333--359},
  year={2011},
  publisher={Springer}
}

@inproceedings{gong2025evolutionary,
  title={Evolutionary large language model for automated feature transformation},
  author={Gong, Nanxu and Reddy, Chandan K and Ying, Wangyang and Chen, Haifeng and Fu, Yanjie},
  booktitle={Proceedings of the AAAI conference on artificial intelligence},
  volume={39},
  number={16},
  pages={16844--16852},
  year={2025}
}

@article{nam2024optimized,
  title={Optimized feature generation for tabular data via llms with decision tree reasoning},
  author={Nam, Jaehyun and Kim, Kyuyoung and Oh, Seunghyuk and Tack, Jihoon and Kim, Jaehyung and Shin, Jinwoo},
  journal={Advances in Neural Information Processing Systems},
  volume={37},
  pages={92352--92380},
  year={2024}
}

@article{zhang2013review,
  title={A review on multi-label learning algorithms},
  author={Zhang, Min-Ling and Zhou, Zhi-Hua},
  journal={IEEE transactions on knowledge and data engineering},
  volume={26},
  number={8},
  pages={1819--1837},
  year={2013},
  publisher={IEEE}
}

@article{zhang2007ml,
  title={ML-KNN: A lazy learning approach to multi-label learning},
  author={Zhang, Min-Ling and Zhou, Zhi-Hua},
  journal={Pattern recognition},
  volume={40},
  number={7},
  pages={2038--2048},
  year={2007},
  publisher={Elsevier}
}

@article{mann2020language,
  title={Language models are few-shot learners},
  author={Mann, Ben and Ryder, Nick and Subbiah, Melanie and Kaplan, J and Dhariwal, P and Neelakantan, A and Shyam, P and Sastry, G and Askell, A and Agarwal, S and others},
  journal={arXiv preprint arXiv:2005.14165},
  volume={1},
  number={3},
  pages={3},
  year={2020}
}

@article{zhao2023survey,
  title={A survey of large language models},
  author={Zhao, Wayne Xin and Zhou, Kun and Li, Junyi and Tang, Tianyi and Wang, Xiaolei and Hou, Yupeng and Min, Yingqian and Zhang, Beichen and Zhang, Junjie and Dong, Zican and others},
  journal={arXiv preprint arXiv:2303.18223},
  volume={1},
  number={2},
  year={2023}
}

@article{dinh2022lift,
  title={Lift: Language-interfaced fine-tuning for non-language machine learning tasks},
  author={Dinh, Tuan and Zeng, Yuchen and Zhang, Ruisu and Lin, Ziqian and Gira, Michael and Rajput, Shashank and Sohn, Jy-yong and Papailiopoulos, Dimitris and Lee, Kangwook},
  journal={Advances in Neural Information Processing Systems},
  volume={35},
  pages={11763--11784},
  year={2022}
}

@article{blei2003latent,
  title={Latent dirichlet allocation},
  author={Blei, David M and Ng, Andrew Y and Jordan, Michael I},
  journal={Journal of machine Learning research},
  volume={3},
  number={Jan},
  pages={993--1022},
  year={2003}
}

@inproceedings{horn2019autofeat,
  title={The autofeat python library for automated feature engineering and selection},
  author={Horn, Franziska and Pack, Robert and Rieger, Michael},
  booktitle={Joint European Conference on Machine Learning and Knowledge Discovery in Databases},
  pages={111--120},
  year={2019},
  organization={Springer}
}

@article{wang2023reinforcement,
  title={Reinforcement-enhanced autoregressive feature transformation: Gradient-steered search in continuous space for postfix expressions},
  author={Wang, Dongjie and Xiao, Meng and Wu, Min and Zhou, Yuanchun and Fu, Yanjie and others},
  journal={Advances in Neural Information Processing Systems},
  volume={36},
  pages={43563--43578},
  year={2023}
}

@inproceedings{liu2015large,
  title={Large margin metric learning for multi-label prediction},
  author={Liu, Weiwei and Tsang, Ivor},
  booktitle={Proceedings of the AAAI Conference on Artificial Intelligence},
  volume={29},
  number={1},
  year={2015}
}

@inproceedings{mao2024learning,
  title={Learning label-specific multiple local metrics for multi-label classification},
  author={Mao, Jun-Xiang and Hang, Jun-Yi and Zhang, Min-Ling},
  booktitle={Proceedings of the 33rd International Joint Conference on Artificial Intelligence},
  pages={4742--4750},
  year={2024}
}

@article{gong2023distributed,
  title={Distributed and joint evidential k-nearest neighbor classification},
  author={Gong, Chaoyu and Demmel, Jim and You, Yang},
  journal={IEEE Transactions on Knowledge and Data Engineering},
  volume={36},
  number={11},
  pages={5972--5985},
  year={2023},
  publisher={IEEE}
}

@article{jia2023learning,
  title={Learning label-specific features for decomposition-based multi-class classification},
  author={Jia, Bin-Bin and Liu, Jun-Ying and Hang, Jun-Yi and Zhang, Min-Ling},
  journal={Frontiers of Computer Science},
  volume={17},
  number={6},
  pages={176348},
  year={2023},
  publisher={Springer}
}

@inproceedings{gouk2016learning,
  title={Learning distance metrics for multi-label classification},
  author={Gouk, Henry and Pfahringer, Bernhard and Cree, Michael},
  booktitle={Asian Conference on machine learning},
  pages={318--333},
  year={2016},
  organization={PMLR}
}

@misc{public2024,
  author = {Public},
  title = {{UCI Dataset Download}},
  year = {2024},
  howpublished= "\url{https://archive.ics.uci.edu/}",
  note= "Accessed: 2024-05-01"
}

@misc{howard2024kaggle,
  author = {Jeremy Howard},
  title = {Kaggle Dataset Download},
  year = {2024},
  note= "Accessed: 2024-05-01",
  howpublished = {\url{https://www.kaggle.com/datasets}}
}

@misc{public2024openml,
  author = {Public},
  year = {2024},
  title = {OpenML Dataset Download},
  note="Accessed: 2024-05-01",
  howpublished = {\url{https://www.openml.org}}
}

@inproceedings{mao2025implicit,
  title={Implicit Relative Labeling-Importance Aware Multi-Label Metric Learning},
  author={Mao, Jun-Xiang and Rui, Yong and Zhang, Min-Ling},
  booktitle={Proceedings of the AAAI Conference on Artificial Intelligence},
  volume={39},
  number={18},
  pages={19414--19422},
  year={2025}
}

@article{hang2024dual,
  title={Dual perspective of label-specific feature learning for multi-label classification},
  author={Hang, Jun-Yi and Zhang, Min-Ling},
  journal={ACM Transactions on Knowledge Discovery from Data},
  volume={19},
  number={1},
  pages={1--30},
  year={2024},
  publisher={ACM New York, NY}
}

@inproceedings{yang2025fast,
  title={Fast Multi-Instance Partial-Label Learning},
  author={Yang, Yin-Fang and Tang, Wei and Zhang, Min-Ling},
  booktitle={Proceedings of the AAAI Conference on Artificial Intelligence},
  volume={39},
  number={21},
  pages={22038--22046},
  year={2025}
}

@inproceedings{zhou2025batch,
  title={Batch Selection for Multi-Label Classification Guided by Uncertainty and Dynamic Label Correlations},
  author={Zhou, Ao and Liu, Bin and Wang, Jin and Tsoumakas, Grigorios},
  booktitle={Proceedings of the AAAI Conference on Artificial Intelligence},
  volume={39},
  number={21},
  pages={22902--22909},
  year={2025}
}

@article{liu2025oversampling,
  title={Oversampling multi-label data based on natural neighbor and label correlation},
  author={Liu, Bin and Zhou, Ao and Wei, Bingkun and Wang, Jin and Tsoumakas, Grigorios},
  journal={Expert Systems with Applications},
  volume={259},
  pages={125257},
  year={2025},
  publisher={Elsevier}
}

@article{hao2024label,
  title={Label generation with consistency on the graph for multi-label feature selection},
  author={Hao, Pingting and Zhang, Ping and Feng, Qi and Gao, Wanfu},
  journal={Information Sciences},
  volume={677},
  pages={120890},
  year={2024},
  publisher={Elsevier}
}

@article{han2025multi,
  title={Multi-label feature selection based on positive sample information weighting},
  author={Han, Qingqi and Shi, Ruikai and Hu, Liang and Gao, Wanfu},
  journal={Knowledge-Based Systems},
  pages={114373},
  year={2025},
  publisher={Elsevier}
}

@article{han2025enhanced,
  title={Enhanced multi-label feature selection considering label-specific relevant information},
  author={Han, Qingqi and Zhao, Zhanpeng and Hu, Liang and Gao, Wanfu},
  journal={Expert Systems with Applications},
  volume={264},
  pages={125819},
  year={2025},
  publisher={Elsevier}
}

@article{hao2024exploring,
  title={Exploring view-specific label relationships for multi-view multi-label feature selection},
  author={Hao, Pingting and Ding, Weiping and Gao, Wanfu and He, Jialong},
  journal={Information Sciences},
  volume={681},
  pages={121215},
  year={2024},
  publisher={Elsevier}
}

@article{hao2024anchor,
  title={Anchor-guided global view reconstruction for multi-view multi-label feature selection},
  author={Hao, Pingting and Liu, Kunpeng and Gao, Wanfu},
  journal={Information Sciences},
  volume={679},
  pages={121124},
  year={2024},
  publisher={Elsevier}
}
\end{document}